\PassOptionsToPackage{svgnames}{xcolor}
\documentclass[electronic]{vgtc}             

\graphicspath{{figures/}{pictures/}{images/}{./}} 

\usepackage{times}                     
\usepackage{amssymb}
\usepackage{tabu}                      
\usepackage{booktabs}                  
\usepackage{lipsum}                    
\usepackage{mwe}                       
\usepackage{amsmath}                    
\usepackage{mathptmx}                  
\usepackage{graphicx} 
\usepackage{epstopdf}
\usepackage{svg}
\usepackage{multirow}
\usepackage [english]{babel}
\usepackage [autostyle, english = american]{csquotes}
 \usepackage[final]{changes} 

\definechangesauthor[name={adder}, color=blue]{+} 
\definechangesauthor[name={deleter}, color=red]{-} 

\MakeOuterQuote{"}

\onlineid{1032}

\vgtccategory{Research}

\vgtcinsertpkg

\title{\textit{EasyREG}: Easy Depth-Based Markerless Registration and Tracking using Augmented Reality Device for Surgical Guidance}

\author{
  Yue Yang\thanks{first author, e-mail: yueyang1@stanford.edu}\\
  \scriptsize Stanford University
  \and
  Christoph Leuze\thanks{e-mail: cleuze@stanford.edu}\\
  \scriptsize Stanford University
  \and
  Brian Hargreaves\thanks{e-mail: bah@stanford.edu}\\
  \scriptsize Stanford University
  \and
  Bruce Daniel\thanks{e-mail: bdaniel@stanford.edu}\\
  \scriptsize Stanford University
  \and
  Fred Baik\thanks{corresponding author, e-mail: fbaik@stanford.edu}\\
  \scriptsize Stanford University
}

\teaser{
  \centering
  \includegraphics[width=\linewidth]{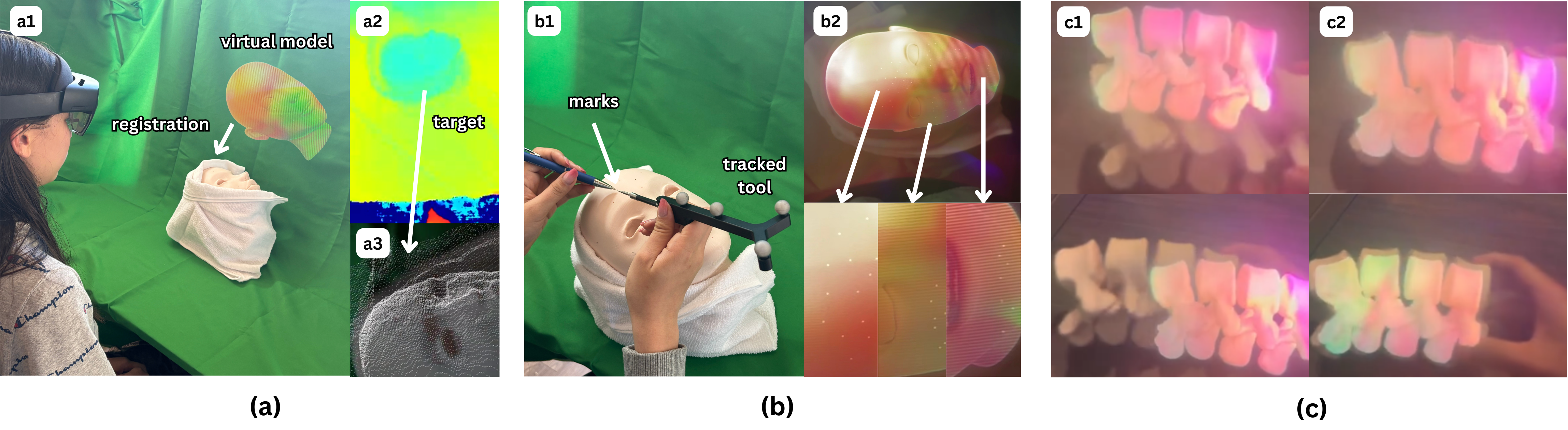}
  \caption{\textit{EasyREG} is a markerless registration and tracking system developed for the Microsoft HoloLens 2, utilizing only the device’s native Time-of-Flight (ToF) depth sensor. \textbf{(a)} In the initial registration phase, the user maintains a stable head position to ensure accurate alignment. \textbf{(b)} A user study was conducted to evaluate registration and tracking performance. In (b1), after executing the registration module, the user uses a tracked tool to mark predefined virtual target points overlaid on the registered 3D hologram. The tracking module then continuously updates the 3D pose of the target in real time as it moves. To assess tracking accuracy, the user repeats the marking task post-motion. \textbf{(c1–c2)} illustrate the hologram dynamically overlaid by the tracking module onto a physical spine model, as captured through the HoloLens lenses.}
  \label{fig:teaser}
  }

\abstract{
     The use of Augmented Reality (AR) devices for surgical guidance has gained increasing traction in the medical field. Traditional registration methods often rely on external fiducial markers to achieve high accuracy and real-time performance. However, these markers introduce cumbersome calibration procedures and can be challenging to deploy in clinical settings. While commercial solutions have attempted real-time markerless tracking using the native RGB cameras of AR devices, their accuracy remains questionable for medical guidance, primarily due to occlusions and significant outliers between the live sensor data and the preoperative target anatomy point cloud derived from MRI or CT scans. In this work, we present a markerless framework that relies only on the depth sensor of AR devices and consists of two modules: a registration module for high-precision, outlier-robust target anatomy localization, and a tracking module for real-time pose estimation. The registration module integrates depth sensor error correction, a human-in-the-loop region filtering technique, and a robust global alignment with curvature-aware feature sampling, followed by local ICP refinement, for markerless alignment of preoperative models with patient anatomy. The tracking module employs a fast and robust registration algorithm that uses the initial pose from the registration module to estimate the target pose in real-time. We comprehensively evaluated the performance of both modules through simulation and real-world measurements. The results indicate that our markerless system achieves superior performance for registration and comparable performance for tracking to industrial solutions. The two-module design makes our system a one-stop solution for surgical procedures where the target anatomy moves or stays static during surgery.
} 


\begin{document}


\firstsection{Introduction}

\maketitle

Augmented Reality (AR) head-mounted displays (HMDs) have demonstrated significant potential in surgical navigation by overlaying critical anatomical and planning information directly onto the surgeon’s field of view. In particular, the Microsoft HoloLens 2 has been increasingly integrated into various surgical workflows, ranging from procedures involving rigid, immobilized anatomy (e.g., spinal and cranial surgeries) to those with more dynamic and mobilized anatomical targets (e.g., knee surgeries) \cite{de2024integrating}. A fundamental requirement for effective AR guidance is the accurate estimation of the three-dimensional pose of the target anatomy (registration) and the ability to dynamically update this pose in real-time (tracking). This process is traditionally achieved using fiducial markers rigidly attached to the patient’s anatomy, which are then tracked using either external optical tracking systems such as OptiTrack or NDI, or through native tracking algorithms embedded within the AR headset \cite{anandan2024surgical, martin2023sttar, birlo2022utility}. Marker-based approaches are widely adopted due to their high accuracy, low latency, and fast response times, making them suitable for a variety of surgical applications \cite{furtado2019comparative,elfring2010assessment}. These methods are also prevalent in robot-assisted surgeries, where markers are typically fixed to bone surfaces to ensure stable tracking \cite{biswas2023recent, lin2021preliminary}.

Despite their advantages, marker-based systems introduce several limitations \cite{canton2024feasibility, hersh2021augmented}. The use of physical markers requires additional procedural steps, including placement, removal, and verification of visibility throughout the procedure. Furthermore, patients are often required to wear these markers during preoperative imaging sessions such as CT or MRI, so that the scanned anatomical models can be registered using the markers as spatial anchors. This imposes constraints on the surgical workflow and creates logistical complications, as the markers must remain fixed in position from the time of imaging through to surgery. Additionally, marker-based systems are limited by the requirement for consistent line-of-sight between the tracking hardware and the markers, which restricts surgical freedom and can be disrupted by occlusions. In response to these limitations, markerless techniques have been developed to exploit the patient’s natural anatomical features for both registration and tracking, thereby eliminating the need for artificial fiducials.

Recent progress in computer vision and depth sensing technologies has enabled the development of markerless methods that support both registration of preoperative models and intraoperative tracking. For example, Groenenberg et al. introduced ARCUS, a depth-based AR navigation system on the HoloLens 2 that performs markerless registration of holograms to patient anatomy under 30 seconds. While the ARCUS system illustrates the feasibility of markerless registration, it lacks dynamic tracking capability and must be re-registered if the target structure moves during surgery. Moreover, although its accuracy is within 10 millimeters, it does not meet surgical standards, which typically require sub-three millimeter accuracy for mid to low-risk surgeries or sub-millimeter level for high-risk scenarios \cite{tondin2022evaluation, tadros2019ductal}. Other efforts have explored image-based tracking methods, which offer lower latency but suffer from even greater error, often exceeding 10 mm, making them unsuitable for precise surgical guidance \cite{suenaga2015vision}. To date, markerless approaches still significantly lag behind marker-based systems in terms of registration accuracy and real-time adaptability, thereby limiting their use in clinical environments. 

This work introduces a novel markerless registration and tracking framework designed for surgical guidance using the HoloLens 2. The system utilizes the built-in Time-of-Flight (ToF) Articulated HAnd Tracking (AHAT) mode depth sensor for accurate, outlier-robust registration, and a separate module for real-time 6-degree-of-freedom (6DoF) tracking of rigid anatomical target. Our approach consists of a two-module architecture. The first module is a robust point cloud registration pipeline that aligns a preoperative 3D model to the intraoperative patient anatomy, even in the presence of substantial outliers. This pipeline uses a coarse-to-fine strategy that begins with a human-in-the-loop step to correct sensor error. It then employs the TEASER++ algorithm for globally optimal registration that is resistant to noise and outliers, followed by a local refinement step using a robust iterative closest point (ICP) method to fine-tune the alignment. The second module addresses real-time tracking through the implementation of a fast ICP algorithm to estimate the pose efficiently for textureless rigid objects at the cost of slightly lower accuracy. The key contributions of this work include: 
\begin{itemize} 
\item A novel markerless registration pipeline for the HoloLens 2 that provides accurate and robust target registration in challenging surgical scenarios.
\item An integration of a tracking algorithm that enables real-time 6DoF tracking of a rigid anatomical target. 
\item An all-in-one solution that eliminates the need for external cameras. 
\item A comprehensive evaluation of the proposed system through simulation and real-world experiments. 
\end{itemize}

To the best of our knowledge, this work presents the first fully markerless method for both registration and tracking that achieves surgical-level accuracy using only the native sensors of the HoloLens 2. Designed specifically for surgical applications, our system integrates depth-based markerless registration and tracking to enable real-time intraoperative guidance on a commercially available AR headset. The registration module can operate independently when the target anatomical structure remains static, while the tracking module can be activated to accommodate intraoperative motion and dynamically adjust the visualization. By eliminating the need for external tracking infrastructure, this work advances the practicality of AR-guided surgery and enhances the flexibility and usability of mixed reality systems in the operating room.

\begin{figure*}[t]
\begin{centering}
    \includegraphics[width=\textwidth]{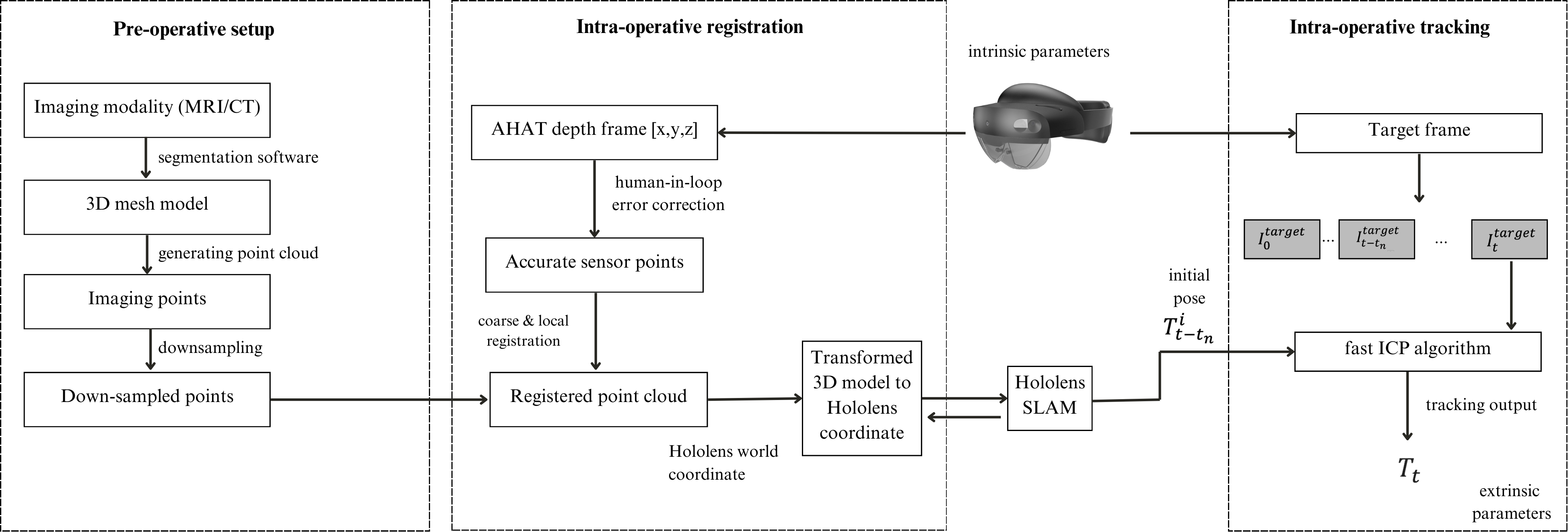}
    \caption{Overview of the proposed system framework. The preoperative setup generates a point cloud representation of the region of interest (ROI) from segmented CT or MRI data. During surgery, this point cloud is aligned to the patient using the registration module, which utilizes depth data from the HoloLens 2’s native ToF sensor to produce the registration transformation. The resulting transformation also serves as the initial pose for the optional tracking module, which continuously updates the model’s position in real-time if the target anatomy moves intraoperatively. The registration module alone is suitable for procedures where the patient and target anatomy remain static, while the tracking module can be activated as needed in dynamic surgical scenarios.}
    \label{fig:eval}
    \end{centering}
\end{figure*}

\section{Related work}
\subsection{Marker-Based AR-Guided Surgery}
AR HMDs, such as the Microsoft HoloLens 2, have been widely investigated for surgical navigation across diverse clinical applications \cite{evans2025augmented, malhotra2023augmented, barcali2022augmented}. Accurate spatial guidance in these systems requires precise knowledge of the patient’s anatomical pose, which is typically achieved through marker-based tracking. Common approaches include the use of optical fiducials—such as ArUco, ChArUco, or QR codes—or passive retro-reflective spheres tracked via infrared cameras \cite{andress2018fly, duan2025localization, cutolo2021device, ma2019augmented}. Commercial navigation systems (e.g., NDI Polaris, OptiTrack) utilize external infrared cameras to track rigidly attached marker frames on the patient or instruments, offering sub-millimeter accuracy but requiring line-of-sight and dedicated external hardware \cite{moreta2022evaluation, anandan2024surgical}.

Recent advances have utilized the native sensing capabilities of the HoloLens 2 for marker-based tracking, demonstrating performance comparable to external tracking systems in applications such as target registration, continuous pose estimation, and real-time surgical tool tracking \cite{martin2023sttar, kihara2024evaluating, li2024navigate}. Despite these advancements, marker-based methods present several practical limitations. Their accuracy can be affected by environmental lighting, occlusion, and the placement of markers \cite{khan2015factors}. More critically, such methods inherently assume a rigid transformation between the marker and the anatomical structure of interest. This assumption often fails in clinical settings, where tissue deformation, patient movement, or intraoperative repositioning may alter the spatial relationship established during preoperative imaging \cite{groenenberg2024feasibility}.

\subsection{Markerless Registration and Tracking Method on AR devices}
Aligning a target 3D anatomical structure, reconstructed from preoperative imaging modalities such as CT or MRI, to the patient without the use of fiducials remains a significant challenge in AR-guided surgery. This difficulty arises from inter-patient anatomical variability, sensor noise, and the presence of substantial outliers in intraoperative data. Broadly, markerless registration and tracking methods in AR surgical systems fall into two categories: depth-based and vision-based approaches. Depth-based methods typically offer higher registration accuracy and robustness but often incur increased computational cost, making them well-suited for static target registration. In contrast, vision-based methods, primarily using RGB cameras, provide faster runtime and are more appropriate for real-time pose estimation and dynamic tracking. Recent advancements in depth-based approaches also enabled fast and real-time object tracking, comparable to vision-based approaches \cite{gao2019filterreg, zhang2021fast}.

Previous researchers have developed depth and vision systems for markerless registration or tracking using AR devices. Kim et al. developed a tablet-based AR system for facial reconstructive surgery that employed an ICP algorithm to align the target anatomy, achieving sub-2 mm registration accuracy \cite{kim2024application}. However, their method was computationally intensive and not integrated with wearable HMDs. More recently, Groenenberg et al. introduced a faster registration technique using the native depth camera of the HoloLens 2, but reported higher registration errors that exceeded 10 mm \cite{groenenberg2024feasibility}. Suenaga et al. demonstrated the feasibility of markerless registration using an external stereo vision system in a surgical setting \cite{suenaga2015vision}. Ahmad et al. proposed a similar stereo vision-based method for automatic dental registration, achieving better alignment accuracy, but with a higher average processing time of approximately 20 seconds \cite{ahmad2024automatic}. Wang et al. also employed stereo vision for oral and maxillofacial surgery, yielding promising results in both accuracy and runtime; however, their system relied on external stereo cameras, which limits its real-world usability \cite{wang2019practical}. Recent efforts have also sought to combine depth and vision. Hu et al. introduced an AI-driven navigation system integrating both modalities, but their approach exhibited translation errors exceeding 5 mm and rotational errors over 5° in a drilling task \cite{hu2024artificial}.

\section{Methods}
We present an end-to-end framework for markerless registration and tracking for AR-guided surgery using the HoloLens 2. The system consists of three main components: 
\begin{itemize}
    \item Preoperative model preparation, where a 3D mesh of the target anatomy is reconstructed from medical imaging and converted into a downsampled point cloud for registration. 
    \item Intraoperative registration module, which robustly aligns the virtual model with the real patient using depth data from the HoloLens 2 ToF sensor.
    \item Intraoperative tracking module, which continuously updates the target's pose using a fast ICP algorithm when the anatomy moves.
\end{itemize}

We validate our system through a comprehensive evaluation involving simulation experiments, a user study using 3D-printed models, and a cadaveric use case. The overall system architecture is shown in \cref{fig:eval}. We provide an overview of the preoperative setup, and a detailed description of the intraoperative registration and tracking modules. 
\subsection{Preoperative Setup}
The first step in markerless AR-guided surgery is obtaining an accurate 3D representation of the target anatomy, typically reconstructed from preoperative CT or MRI scans. Segmentation of the region of interest (ROI) is performed using 3D Slicer. After segmentation, we use the Open3D Python library to convert the mesh into a point cloud and downsample it to 5,000 points to balance computational efficiency and registration accuracy.

The system is implemented using the HoloLens 2 as the AR platform, paired with a high-performance workstation (Intel® Core™ i9-13950HX CPU, NVIDIA GeForce RTX 4060 GPU). The headset communicates with the workstation via a custom TCP protocol over a high-speed (1000+ Mbps) USB tether, enabling real-time data streaming at approximately 30 frames per second.

\subsection{Camera Calibration and Registration}
The goal of the registration module is to align the target anatomy to the HoloLens 2’s coordinate system using depth data from the ToF camera (AHAT). Since the initial camera pose obtained here is later used by a separate tracking module with results displayed in the Hololens 2 coordinate, a precise spatial calibration of AHAT cameras is essential. To investigate the feasibility of using the native front-facing RGB camera instead of the depth sensor for tracking, we further calibrated and registered the RGB camera to AHAT for a device-specific field of view (FoV) analysis. 

\subsubsection{Camera Intrinsic Calibration}
Intrinsic calibration of both the AHAT depth sensor and the front-facing RGB camera was performed independently to account for their differing sensing modalities. Calibration of the RGB camera used a planar checkerboard with 7×5 inner corners and 30 mm square sizes. High-resolution images (3904×2196 pixels) were captured from various positions, angles, and distances to ensure adequate coverage. Feature points were detected and used in a non-linear optimization process to estimate focal lengths, principal point coordinates, and distortion coefficients by minimizing the reprojection error:
\begin{equation}
    \text{RMSE} = \sqrt{\frac{1}{N} \sum_{i=1}^{N} \|x_i - \hat{x}_i\|^2}
\end{equation}
where $x_{i}$ is the detected image point and $\hat{x}_i$ is the reprojected point from the estimated 3D parameters. The AHAT sensor was calibrated using its reflectivity images. A total of 100 frames were collected: 70 for calibration and 30 for evaluation. This procedure yielded reprojection errors of 0.268 ± 0.031 pixels and 0.266 ± 0.050 pixels for the RGB camera, and 0.148 ± 0.053 and 0.162 ± 0.040 pixels for the AHAT sensor.
\subsubsection{RGB–AHAT Extrinsic Calibration}
Following intrinsic calibration, we computed the extrinsic transformation between the RGB and AHAT cameras to align their coordinate systems. This process involved estimating a rigid-body transformation, rotation matrix $\mathbf{R} \in SO(3)$ and translation vector $\mathbf{t} \in \mathbb{R}^3$, that maps 3D points from the RGB camera space to the AHAT space.

Using the previously captured RGB images and known camera intrinsics, 3D corner points on the checkerboard were triangulated and paired with their corresponding detections in the AHAT reflectivity images. The transformation was computed by minimizing the mean squared error (MSE) between corresponding 3D point sets \{$r_i$\} (RGB) and \{$a_i$\} (AHAT):
\begin{equation}
    \min_{\mathbf{R}, \mathbf{t}} \sum_{i=1}^{N} \left\| \mathbf{a}_i - (\mathbf{R} \mathbf{r}_i + \mathbf{t}) \right\|^2
\end{equation}

To assess extrinsic calibration accuracy, 3D points from the RGB camera were transformed into AHAT space, projected onto the AHAT image plane, and compared to ground-truth AHAT detections. The reprojection RMSE was below 1 pixel (0.832), indicating high spatial alignment fidelity. Using similar approach by \cite{martin2023sttar}, we measured the sensor FoV on our device, resulting in $40.3^\circ \times 64.8^\circ$ (RGB) and $126.8^\circ \times 126.6^\circ$ (AHAT) respectively. The significant larger FoV of the AHAT camera makes it more suitable for surgical registration. 

\subsection{Intra-operative Registration}
The registration module is outlined in \cref{fig:eval}. Raw depth data from the HoloLens 2’s AHAT sensor are acquired via the device’s research mode and streamed in real-time to the external workstation described previously. To ensure accurate alignment, we first evaluated the depth accuracy of the AHAT camera on different body regions and introduced a region-specific correction method to mitigate sensor noise.

Following this, the pre-generated point cloud of the target anatomical model (obtained during preoperative setup) is registered to the intraoperatively acquired depth data. This process involves a coarse correspondence-based registration algorithm that is robust to outliers, followed by a local point-to-plane ICP refinement step to improve registration accuracy and convergence near the final pose.
\subsubsection{Evaluation of AHAT depth sensor accuracy}
\label{sec:calibration}
To enable accurate registration, it is critical that the point cloud acquired from the HoloLens 2's ToF AHAT depth sensor accurately reflects the geometry of the target anatomy. Prior studies have shown that ToF sensors exhibit varying degrees of depth error depending on the surface material properties \cite{li2024evd, groenenberg2024feasibility}. Specifically, retro-reflective materials, known for their strong infrared reflectivity, can achieve sub-millimeter accuracy when combined with Kalman filtering for marker tracking \cite{martin2023sttar}. In contrast, common 3D printing materials have been reported to result in depth errors exceeding 10 mm due to poor infrared response.

\begin{figure*}[t]
\begin{centering}
    \includegraphics[width=\textwidth]{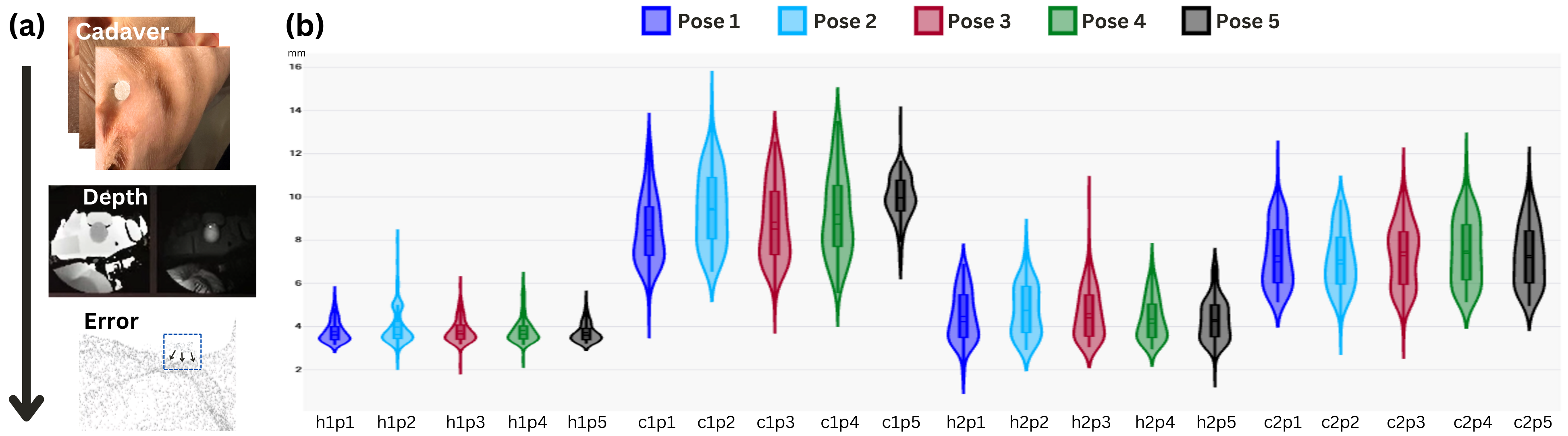}
    \caption{AHAT depth sensor data collection and error distribution. \textbf{(a)} Workflow for cadaveric data collection using the HoloLens 2 AHAT depth sensor. Retroreflective markers were applied to multiple anatomical regions, and depth data were captured from multiple poses to assess sensor accuracy. \textbf{(b)} Distribution of depth error for each data point across all evaluated conditions. Labels such as $h1p1$ refer to head marker 1 at pose 1, and $c1p1$ refers to chest marker 1 at pose 1.}
    \label{fig:sensor}
    \end{centering}
\end{figure*}

To assess whether similar error characteristics are present when scanning human tissue, we conducted a quantitative evaluation of the relative depth error on cadaveric skin under surgical light, using retro-reflective tape as a ground truth reference for surface depth. Four pieces of retro-reflective tape were affixed directly to a cadaver at anatomically distinct regions (each at least 200mm apart) with relatively flat skin surfaces: chest (2 locations: $c1$–$c2$), head (2 locations: $h1$-$h2$), see \cref{fig:sensor} (a). These regions were chosen to cover different anatomical zones while ensuring sufficient visibility for depth sensing.

The HoloLens 2 was worn by an operator, who recorded depth data from 25 different camera poses while maintaining the target at arm's distance, with minor head movement. This also ensures that all four retro-reflective markers were visible in each capture. For analysis, five poses were randomly selected from the 25. In one case, a pose was excluded because not all four markers were captured during data analysis, and an alternative pose was randomly chosen to maintain consistency across samples. To extract reference depth data, the retro-reflective regions were first segmented based on their high intensity values in the infrared reflectivity channel. The corresponding points in the depth map, indexed by the set $R=\{1,2,\ldots,R\}$, were then reprojected into 3D space using the camera intrinsic and transformation matrix $K^{-1}$, yielding spatial coordinates $X_r = (x_r,y_r,z_r) \in \mathbb{R}^3$ for $r \in R$. These $R$ points were used to estimate a reference plane $\Pi : Ax + By + Cz + D = 0$ via least-squares fitting. To assess local depth accuracy, we extracted 3D points $\mathbf{P}_k = (x_k, y_k, z_k)$ from the surrounding cadaver skin surface within a 20 mm radius from the reflective tape denoted as the neighborhood set $S = \{1, 2, \dots, S\}$. Assuming these points ideally lie on the same reference plane, we projected each $\mathbf{P}_k \in S$ orthogonally onto $\Pi$:
\begin{equation}
    \mathbf{P}'_k = \mathbf{P}_k - \frac{Ax_k + By_k + Cz_k + D}{A^2 + B^2 + C^2} \cdot (A, B, C) \label{eq:3}
\end{equation}
The pointwise depth error $\delta_k$ was then calculated as the Euclidean distance between the measured point and its planar projection:
\begin{equation}
    \delta_k = \|\mathbf{P}_k - \mathbf{P}'_k\|
\end{equation}
The angle of incidence $\theta$ and depth between the HoloLens 2 and the cadaver skin surface varied naturally between each retroreflective tape region, as the tape placements were spatially distributed across distinct anatomical regions ($>$200mm apart). \cref{fig:sensor} presents the distribution of calculated depth error for each data point under varying conditions.

A Shapiro-Wilk test indicated significant departures from normality for all four anatomical regions on the cadaver (all $p<.001$). As the normality assumption was violated, a non-parametric Friedman test was performed, revealing statistically significant differences in depth error across the regions ($p<.001$). A subsequent post-hoc Nemenyi test confirmed significant pairwise differences between all groups ($p<.001$), suggesting that depth error varies significantly across different anatomical regions. These findings imply that systematic depth error correction must be tailored to the specific anatomical region being registered, rather than applying a uniform transformation across the entire point cloud. A global correction approach would be insufficient to eliminate localized sensor bias, emphasizing the need for case-by-case error correction in AR-guided surgery workflows. Since error correction needs to be done frequently, our proposed error correction method balances convenience and effectiveness (described in \cref{sec:errorcorrection}). 

Furthermore, we assessed the consistency of depth error across different camera poses for each anatomical region. When comparing five different poses at each target location (e.g., $h1p1$ to $h1p5$), the Shapiro-Wilk test again indicated non-normal distributions for all cases. Friedman tests for regions $h1$, $h3$, and $h4$ showed no statistically significant differences across poses ($h1: p=0.63$; $h3: p=0.38$; $h4: p=0.17$), suggesting that minor pose variations during data acquisition do not significantly affect depth error, and the measurements are stable and reproducible. However, the $c1$ region exhibited a statistically significant result ($p<.001$). Post-hoc pairwise Nemenyi tests revealed significant differences for the following pose pairs: $c_1p_1 \text{ vs. } c_1p_5 \,(Q = 10.8 > Q_{\text{crit}} = 3.86)$; $c_1p_2 \text{ vs. } c_1p_5 \,(Q = 4.6 > Q_{\text{crit}} = 3.86)$; $c_1p_3 \text{ vs. } c_1p_5 \,(Q = 7.5 > Q_{\text{crit}} = 3.86)$

These differences are driven primarily by outliers in the $c1p5$ group. All other comparisons remained statistically insignificant, suggesting that the depth error is generally stable across small head pose changes.

\subsubsection{Region-Specific Depth Sensor Error Correction}
\label{sec:errorcorrection}
To correct the previous described depth sensor error, researchers have proposed a fiducial marker-based approach \cite{li2024evd}. Here, we introduce a region-specific and convenient method without placing additional markers on the patient surface for error correction. This is tailored for surgical purposes as one target region is typically defined prior to surgery for each patient, and accuracy for this region is the key consideration. 

Consider a target region $M$ defined as a circular area with a 70 $mm$ radius. We first obtain a set of ground truth points by using an AHAT sensor–tracked tool (by \cite{martin2023sttar}) to map the true contour of $M$. Let \begin{equation}
    L = \{L_i \in \mathbb{R}^3 \mid i = 1, 2, \dots, n\}
\end{equation} denote these ground truth points, which inherently lie within $M$.
To align the sensor’s measurements with these ground truth points, we need to extract the corresponding sensor-measured points from the depth data. For each ground truth point $L_i$, we identify the corresponding sensor-measured point $P_i$ as the nearest neighbor within $P_M$ and the paired set:
\begin{equation}
    P_i = \arg\min_{P \in {P}_{all}} \|P - L_i\|, \quad \{(L_i, P_i) \mid i = 1, 2, \dots, n\}
\end{equation}
Our goal is to find the rotation matrix \( \mathbf{R} \in \mathbb{R}^{3 \times 3} \) and translation vector \( \mathbf{t} \in \mathbb{R}^3 \) such that the transformed sensor points
\begin{equation}
    P_i^{\text{corr}} = \mathbf{R} \, P_i + \mathbf{t}
\end{equation}
are as close as possible to the corresponding ground truth points \( L_i \) in a least-squares sense. This is achieved by minimizing the following objective and then computing the centroids of the paired point sets:
\begin{align}
    \min_{\mathbf{R},\,\mathbf{t}} \; \frac{1}{2} \sum_{i=1}^{n} \left\| L_i - \left( \mathbf{R} \, P_i + \mathbf{t} \right) \right\|^2 \label{eq:cost} \\
    \bar{P} = \frac{1}{n} \sum_{i=1}^{n} P_i, \quad \bar{L} = \frac{1}{n} \sum_{i=1}^{n} L_i
\end{align}

We calculate the optimal translation that aligns these centroids and center the point sets for rotation matrix calculation
\begin{equation}
    \mathbf{t} = \bar{L} - \bar{P}
\end{equation}
\begin{align}
    P_i' &= P_i - \bar{P}, \quad L_i' = L_i - \bar{L}
\end{align}
From the covariance matrix, we perform the singular value decomposition (SVD) of \( \mathbf{H} \), and compute the optimal rotation matrix \( \mathbf{R} \):
\begin{equation}
    \mathbf{H} = \sum_{i=1}^{n} P_i' (L_i')^\top
\end{equation}
\begin{equation}
    \mathbf{H} = \mathbf{U} \, \boldsymbol{\Sigma} \, \mathbf{V}^\top
\end{equation}
\begin{equation}
    \mathbf{R} = \mathbf{V} \, \mathbf{U}^\top
\end{equation}
If \( \det(\mathbf{R}) < 0 \), the algorithm adjusts by flipping the sign of the last column of \( \mathbf{V} \) to ensure that \( \det(\mathbf{R}) = 1 \).
Using the proposed approach, we performed error correction for each of the pre-described region ($h1$-$h2$, $c1$-$c2$). Using the same error measurements described in \cref{eq:3}, we reduced median relative error by 82\% for $h1$ (0.72 mm, targeting $h1$), 85\% for $h2$ (0.69 mm, targeting $h2$), 77\% for $c1$ (0.91 mm, targeting $c1$), and 88\% for $c2$ (0.83 mm, targeting $c2$). When targeting each region, the errors for other regions are also corrected with $>$50\% error-reduction rate across measurements, making the resulting point cloud dataset suitable for registration. 

\subsubsection{Coarse Registration via Modified TEASER++ with Curvature-Aware Sampling}
After sensor error correction, the intraoperative scene and preoperative target anatomy are represented by
\begin{align}
\mathcal{P}_s = \{ \mathbf{p}_i \in \mathbb{R}^3 \mid i = 1,\dots,N_s \} \\
\mathcal{P}_t = \{ \mathbf{q}_j \in \mathbb{R}^3 \mid j = 1,\dots,N_t \}
\end{align}

Traditional methods generate feature correspondences uniformly using FPFH. In the context of anatomical registration, however, high-curvature regions (e.g., ridges and contours) are more likely to coincide with salient anatomical landmarks. To preferentially select such regions, we compute a curvature value \( k_i \) for each point \( \mathbf{p}_i \) using local differential geometry, and define a curvature-aware sampling probability:
\begin{equation}
    P(\mathbf{p}_i) \propto \frac{k_i}{\sum_{j=1}^{N_s} k_j}
\end{equation}

This sampling approach biases feature extraction toward landmark-rich areas, enhancing the discriminative quality of the FPFH descriptors, denoted by \( \phi(\mathbf{p}_i) \) for points in \(\mathcal{P}_s\) and \( \phi(\mathbf{q}_j) \) for points in \(\mathcal{P}_t\).

Putative correspondences are then generated by a nearest-neighbor search in the descriptor space:
\begin{equation}
\mathcal{C} = \{ (i,j) \mid \|\phi(\mathbf{p}_i) - \phi(\mathbf{q}_j)\| < \tau \}
\end{equation}
where \(\tau\) is a fixed matching threshold. Each correspondence is further assigned a weight \( w_{ij} \) that incorporates both descriptor similarity and the curvature-based importance.

Due to the high outlier rate typical in keypoint matching for anatomical data, we adopt a robust truncated least squares (TLS) formulation from TEASER++. Our model assumes that the transformation between the two point clouds adheres to
\begin{equation}
    \mathbf{q}_j = \kappa\, \mathbf{R}\, \mathbf{p}_i + \mathbf{t} + \mathbf{o}_i + \boldsymbol{\varepsilon}_i
\end{equation}
where:\( \kappa > 0 \) is the unknown global scale factor;
     \(\mathbf{R} \in SO(3)\) is the rotation matrix;
     \(\mathbf{t} \in \mathbb{R}^3\) is the translation vector;
    \(\mathbf{o}_i\) is the outlier term (zero for inliers);
     \(\boldsymbol{\varepsilon}_i\) denotes bounded measurement noise.

The robust registration problem is then formulated as
\begin{equation}
    \min_{\kappa > 0,\; \mathbf{R}\in SO(3),\; \mathbf{t}\in\mathbb{R}^3} \; \sum_{(i,j)\in \mathcal{C}} w_{ij}\; \min\Big\{ \Big\|\mathbf{q}_j - \big(\kappa\, \mathbf{R}\,\mathbf{p}_i + \mathbf{t}\big)\Big\|^2,\; \epsilon^2 \Big\}
\end{equation}
where \(\epsilon\) is a fixed truncation threshold chosen based on the expected noise level.

To decouple the estimation of scale, rotation, and translation, we form translation-invariant measurements. For any two correspondences \((\mathbf{p}_i,\mathbf{q}_j)\) and \((\mathbf{p}_k,\mathbf{q}_\ell)\), define the pairwise differences
\begin{align}
    \Delta \mathbf{p}_{ik} = \mathbf{p}_i - \mathbf{p}_k,\qquad
\Delta \mathbf{q}_{j\ell} = \mathbf{q}_j - \mathbf{q}_\ell
\end{align}

Assuming the correspondences are inliers, the transformation implies
\begin{equation}
    \Delta \mathbf{q}_{j\ell} = \kappa\, \mathbf{R}\,\Delta \mathbf{p}_{ik}
\end{equation}
\begin{equation}
    \|\Delta \mathbf{q}_{j\ell}\| = \kappa\, \|\Delta \mathbf{p}_{ik}\|
\end{equation}
We then define the scale measurement for the pair and estimate the global scale \( \hat{\kappa} \) via a robust TLS formulation:
\begin{equation}
\kappa_{ik} = \frac{\|\Delta \mathbf{q}_{j\ell}\|}{\|\Delta \mathbf{p}_{ik}\|}
\end{equation}
and estimate the global scale \( \hat{\kappa} \) via a robust TLS formulation:
\begin{equation}
\hat{\kappa} = \arg\min_{\kappa > 0} \; \sum_{(i,k)\in \mathcal{E}} w_{ik}\; \min\left\{ \frac{(\kappa - \kappa_{ik})^2}{\alpha_{ik}^2},\; c^2 \right\}
\end{equation}
where \( \mathcal{E} \subset \mathcal{C} \) is an appropriate set of pairwise correspondences, \(\alpha_{ik}\) are the noise bounds for the corresponding measurements, and \( c^2 \) is a normalization constant (typically \( c^2 = 1 \)).

With the scale estimate \( \hat{\kappa} \) obtained, we address the rotation subproblem. The rotation is estimated from the translation-invariant relation
\begin{equation}
    \Delta \mathbf{q}_{j\ell} = \hat{\kappa}\, \mathbf{R}\,\Delta \mathbf{p}_{ik} + \text{noise}
\end{equation}

Thus, we solve
\begin{equation}
    \hat{\mathbf{R}} = \arg\min_{\mathbf{R} \in SO(3)} \; \sum_{(i,k)\in \mathcal{E}} w_{ik}\; \min\left\{ \frac{\|\Delta \mathbf{q}_{j\ell} - \hat{\kappa}\, \mathbf{R}\,\Delta \mathbf{p}_{ik}\|^2}{\delta_{ik}^2},\; c^2 \right\}
\end{equation}
where \(\delta_{ik}\) captures the noise level for the measurement pair. This robust rotation estimation utilizes the TEASER++ framework (via a heuristic based on graduated nonconvexity).

Finally, the translation subproblem is solved by aligning the centroids of the weighted inlier correspondences. In practice, the translation is estimated by solving the following component-wise TLS problems:
\begin{align}
    \hat{t}_l = \arg\min_{t_l \in \mathbb{R}} \; \sum_{(i,j) \in \mathcal{C}} w_{ij}\; \min\left\{ \frac{\Big| \Big[ \mathbf{q}_j - (\hat{\kappa}\,\hat{\mathbf{R}}\,\mathbf{p}_i + \mathbf{t})\Big]_l \Big|^2}{\beta_i^2},\; c^2 \right\}
\end{align}
where $l$=1,2,3 denotes the \(l\)th component and \(\beta_i\) represents the noise bound for translation.

The resulting coarse registration estimate \( (\hat{\kappa}, \hat{\mathbf{R}}, \hat{\mathbf{t}}) \) is subsequently refined using a point-to-plane ICP algorithm to achieve the precise alignment required for AR-guided surgical navigation.

We further conducted a preliminary evaluation of the coarse registration algorithm using a cadaver model to assess the success of global registration, with error correction targeting the forehead region. Detailed quantitative results and comprehensive evaluations are presented in \cref{section:result}.

\subsubsection{Local Refinement through ICP}
\label{sec:customdata}
\begin{figure}[tbh]
\begin{centering}
    \includegraphics[width=\linewidth]{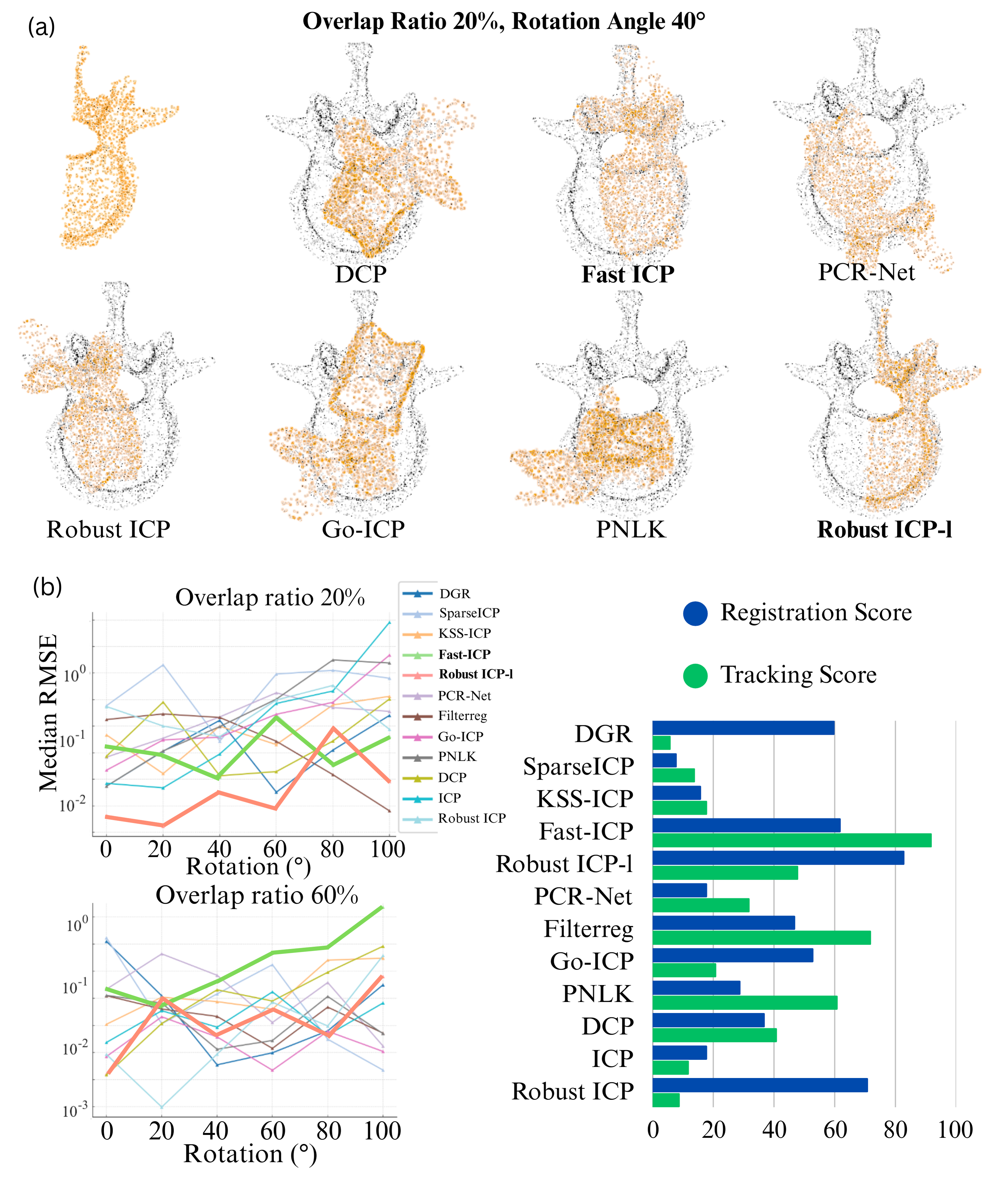}
    \caption{Comparison of 12 registration algorithms on the NIH3D dataset under challenging conditions (0.33 Gaussian noise and 50\% partial data selection). \textbf{(a)} Qualitative registration results at an overlap ratio of 20\% and a rotation angle of 40°, shown for a representative subset of methods. Two selected methods are bolded. \textbf{(b)} Quantitative comparison across all 12 methods under each condition, with computed registration and tracking scores reported.}
    \label{fig:data}
    \end{centering}
\end{figure}

The coarse registration result is refined using a local ICP algorithm to achieve the high accuracy required for AR-guided surgical procedures. While deep learning–based methods have demonstrated strong performance on certain benchmark datasets, they often struggle to generalize to anatomical structures derived from real MRI/CT scan data \cite{wang2020unsupervised, aoki2019pointnetlk, wang2019prnet, sarode2019pcrnet, wang2019deep, choy2020deep}. Traditional ICP algorithms have seen substantial improvements over the past decade, but their performance must still be carefully evaluated in medical imaging contexts \cite{zhang2021fast, low2004linear, yang2015globally, gao2019filterreg, bouaziz2013sparse, lv2023kss, jian2010robust}. We conducted a simulation experiment using real MRI/CT data from NIH3D database to evaluate 12 ICP variants (See \cref{fig:data}: DGR, SpareseICP, KSS-ICP, Fast-ICP, Robust ICP-l, PCR-Net, Filterreg, Go-ICP, PNLK, DCP, ICP, Robust ICP) \cite{NIH3D}.

Given that registration accuracy is critical for surgical applications, we assign a higher weight to the accuracy metric, expressed as the median RMSE, relative to runtime in our composite scoring method. Specifically, for each model \(m\), let: \(E_{\text{RMSE}, m}\) denote the median RMSE, \(E_{\max}\) denote the maximum of the median RMSE across all models, \(T_{m}\) denote the median runtime, and \(T_{\max}\) denote the maximum of the median runtimes observed among the models. The composite score \(S_m\) for model \(m\) is defined as:
\begin{equation}
\label{eq:eq29}
    S_m = 100 \times \left[ 1 - \left( \lambda \cdot \frac{E_{\text{RMSE}, m}}{E_{\max}} + (1 - \lambda) \cdot \frac{T_{m}}{T_{\max}} \right) \right],
\end{equation}
where \(\lambda\) is the weight factor for accuracy. In our experiments, we set \(\lambda = 0.7\) to place a higher emphasis on accuracy relative to runtime. With this formulation, a model with the lowest normalized median RMSE and runtime receives a score of 100 (indicating the best performance), whereas a model with the worst observed performance receives a score of 0. We describe and report our results in \cref{section:sim} and \cref{fig:data}. 

Among the tested methods, the robust point-to-plane ICP algorithm proposed by Zhang et al. achieved the highest accuracy and was subsequently integrated into our registration framework \cite{zhang2021fast}. The output of the coarse registration was used as input for fine registration. An Axis-Aligned Bounding Box (AABB) around globally registered target anatomy was used to filter captured scene points for better efficiency, and point cloud normals are calculated using PCL library \cite{PCL}. 

\subsection{Intra-operative Tracking}
Using the method from \cref{eq:eq29}, we set \(\lambda = 0.15\) to prioritize run-time speed and evaluated the performance of various algorithms by assigning each a tracking score. To enable real-time pose updates of the target anatomy during mobilized surgery, we utilize the Fast-ICP mode of the algorithm \cite{zhang2021fast}. Let \( I_t^{\text{target}} \) denote the sensor frame at time \( t \), and our goal is to estimate the 6-DoF pose \( \mathbf{T}_t \in SE(3) \). The tracking module depends on a registration module, which provides an initial pose \( \mathbf{T}_{t - t_n}^i \) from a previous frame \( I_{t - t_n}^{\text{target}} \), where \( t_n \) represents the registration latency. Although the registration is not performed in real-time, it is assumed valid because the user's head remains still while wearing Hololens 2 during this initialization step.

Due to the continuous nature of object motion, we utilize temporal coherence by using the most recent valid pose for initialization. The pose initialization logic for time \( t \), denoted \( \mathbf{T}_t^{\text{init}} \), is defined as:
\[
\mathbf{T}_t^{\text{init}} = 
\begin{cases}
\mathbf{T}_{t-1}, & \text{if tracking succeeded at } t-1 \\
\mathbf{T}_{t-k}, & \text{if tracking failed at } t-1 \text{ but succeeded at } t-k < t \\
\mathbf{T}_{t - t_n}^i, & \text{if no valid tracked pose exists} \\
\mathbf{0}, & \text{if neither tracking nor registration has succeeded}
\end{cases}
\]

Thus, the final estimated pose at time \( t \) is computed by the Fast-ICP tracker as:
\[
\mathbf{T}_t = \text{FastICP}(I_t^{\text{target}}, \mathbf{T}_t^{\text{init}})
\]

\section{Results}
\label{section:result}
\subsection{Simulation Experiment}
\label{section:sim}

\begin{figure*}[tbh]
\begin{centering}
    \includegraphics[width=\linewidth]{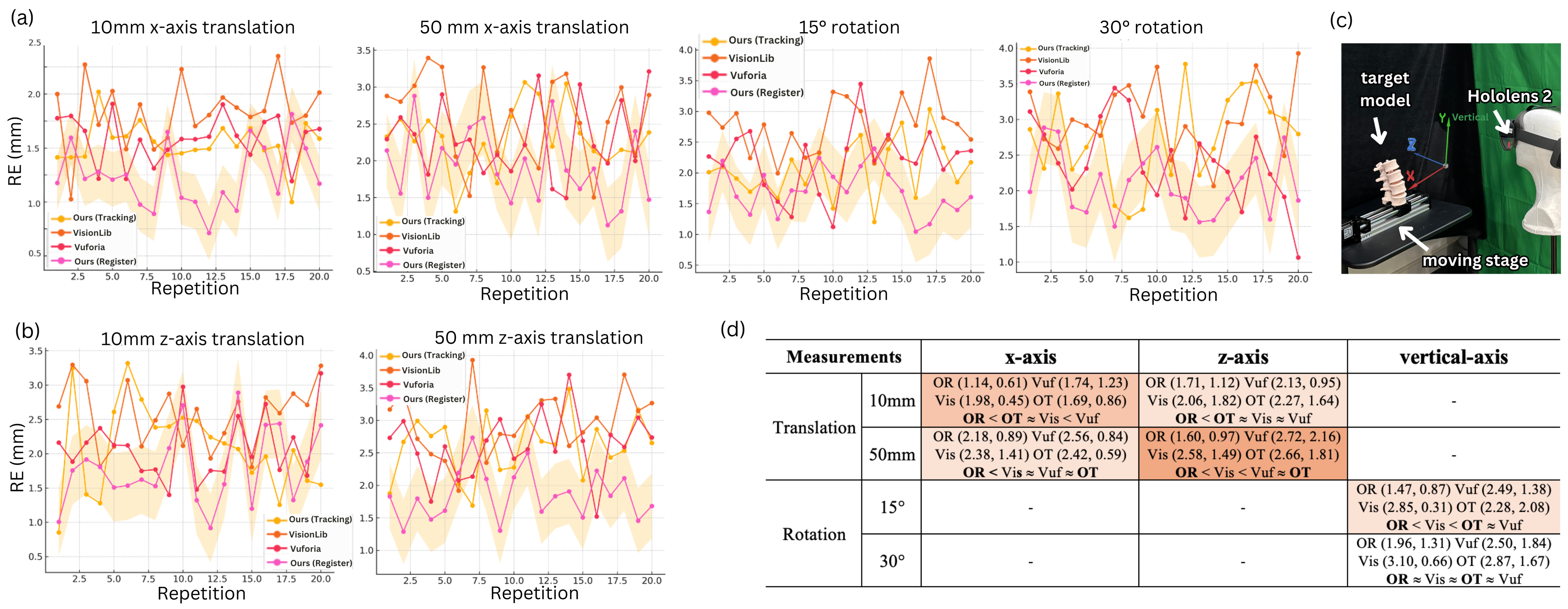}
    \caption{Comparison of our method's registration and tracking module with two industry solutions for markerless registration or tracking on HoloLens 2. Our registration module is visualized in pink with a light yellow band representing IQR, while the tracking module is visualized in yellow. Other methods are shown in orange or red color. \textbf{(a)} Median relative error (RE) across measurement conditions. \textbf{(b)} Independent measurements on other conditions. \textbf{(c)} Experiment setup. \textbf{(d)} Collected translation and rotation RE measurements across conditions for Our Registration module (OR), Our Tracking module (OT), Vuforia (Vuf), and VisionLib (Vis). Data are reported in (Median, IQR) format. Significant differences are highlighted as follows: light yellow for $p < .05$, yellow for $p < .01$, and dark yellow for $p < .001$.}
    \label{fig:stage}
    \end{centering}
\end{figure*}

We present evaluation results of the 12 local registration algorithms in \cref{fig:data}. Six representative target anatomies from the NIH3D dataset (real 3D reconstructions derived from MRI/CT scans) are used for analysis. For each anatomy, we prepare a global point set (downsampled to 8K points) and a 50\% partially selected version with 0.33 Gaussian noise added to simulate realistic depth sensor data. A test set is then generated with overlap ratios of 20\% and 60\%, and rotation angles ranging from 0° to 100° between target points and simulated sensor points. A total of 12 algorithms are evaluated on this dataset. Qualitative results on a spine model are shown in \cref{fig:data} (a). Median RMSE values for each algorithm, aggregated across the six anatomies, are presented under each test condition. Using median runtime data, we compute both registration and tracking scores for each overlap setting and report their average in \cref{fig:data} (b). Based on these evaluations, the Robust ICP-L algorithm and Fast ICP are selected for local registration and tracking respectively, due to their better performance.

To evaluate the accuracy and stability of our proposed markerless registration and tracking framework, we conducted comparative experiments against two industry-standard solutions, Vuforia (SDK Version 6.5.22) and VisionLib (Version 3.2, Trial License), both of which operate on the HoloLens 2 using the front-facing RGB camera and proprietary tracking algorithms. We used a 3-axis electronically controlled translation stage (accuracy: 0.01 mm) and a rotational platform (accuracy: 0.1$^{\circ}$) to provide precise ground truth motion. A 3D-printed human spine model was affixed to the stage, and ToF sensor errors were corrected using our proposed error compensation method. The translation stage was controlled using an Arduino Uno Rev3. During translation experiments, the stage was moved along the x-axis by 10 mm increments. After each movement, the stage was held idle for 30 seconds to allow stabilization and accuracy measurements. At each stage position, 50 poses were recorded. For each movement, 500 relative distance values were randomly sampled from adjacent positions. The difference between these measurements and the known translation (e.g., 10 mm or 50 mm) was calculated to evaluate tracking accuracy. We report the median and interquartile range (IQR) of the calculated relative error for each adjacent pose under four independently tested methods. 

This process was repeated for: 1) additional 10 mm translations to maintain similar viewing angles and eliminate the need for error re-compensation (20 back-and-forth repetitions); 2) 50 mm translations along the x- and z-axes; 3) rotational motions of 15$^{\circ}$ and 30$^{\circ}$ about the vertical axis. The relative error was computed as the absolute difference between the measured translation/rotation and the ground truth. This experimental setup allowed for a rigorous, repeatable comparison of performance between our methods and existing RGB-based industry alternatives under controlled motion and varying magnitudes.

The Shapiro–Wilk test indicated that the data were not normally distributed ($p < .05$). Consequently, we employed non-parametric statistical methods. The Kruskal–Wallis test was used to assess differences between independent measurements across all methods. For pairwise comparisons under each translation or rotation condition, Dunn’s post-hoc tests with Bonferroni correction were applied. Additionally, Mann–Whitney U tests were conducted to compare aggregated data from the $x$-axis and $z$-axis translation groups, which showed insignificant result. As shown in \cref{fig:stage}, all reported measurements are expressed as median and interquartile range (Median, IQR), with statistically significant $p$-values clearly highlighted. Our subsequent analyses focus on insights learned from statistically significant results.

Our registration module achieved the highest overall accuracy across both translation and rotation measurements. Notably, it demonstrated significantly better performance than competing methods during the 50 mm $z$-axis translation condition. This superior accuracy can be attributed to several factors. First, our error correction method is region-specific, and the 50 mm translation likely remains within the corrected region, yielding a highly accurate source point cloud. Second, the target model, a 3D-printed spine with intricate anatomical contours, benefits from our curvature-aware sampling strategy. This strategy emphasizes contour correspondence, which enhances performance compared to standard TEASER++ using uniform FPFH-based correspondences. Third, the local registration step in our method utilizes a robust ICP-l algorithm, which has demonstrated superior registration performance relative to other algorithms, making it particularly suitable for anatomical structure registration tasks. As a result of this accurate initial registration, our tracking module, initialized by this pose, achieves performance comparable to existing industry solutions.

However, our registration module did not outperform other methods during repeated measurements involving high-degree vertical-axis rotations. Several factors may contribute this observation. First, FPFH descriptors are not fully rotation-invariant. TEASER++ performance degrades under large rotations due to unreliable local descriptors, which increases the likelihood of false correspondences. This compromises the effectiveness of the algorithm’s inlier pruning via maximum clique selection and undermines its rotation estimation. Second, as illustrated in \cref{fig:data}, the median RMSE of our integrated local registration algorithm (Robust ICP-l) increases with greater rotational differences between point clouds. Therefore, if TEASER++ outputs a large relative rotation, the final registration accuracy may deteriorate. Third, since depth sensor errors were corrected only once during the initial setup in our experiment, a 30° rotation may move the target region outside the corrected volume. Given that statistically significant differences exist across region pairs (see \cref{fig:sensor}), re-conducting sensor corrections for different rotational configurations may further reduce error and improve accuracy. 

To evaluate the runtime of our tracking module, we conducted a separate experiment in which time stamps were recorded for each processed frame. The median runtime was 210 ms, corresponding to an effective frame rate of approximately 5 FPS. To enhance the smoothness of the user experience during visualization, we implemented a between-frame interpolation strategy, which increased the visual refresh rate to approximately 30 FPS.

\begin{figure}[tbh]
\begin{centering}
    \includegraphics[width = \linewidth]{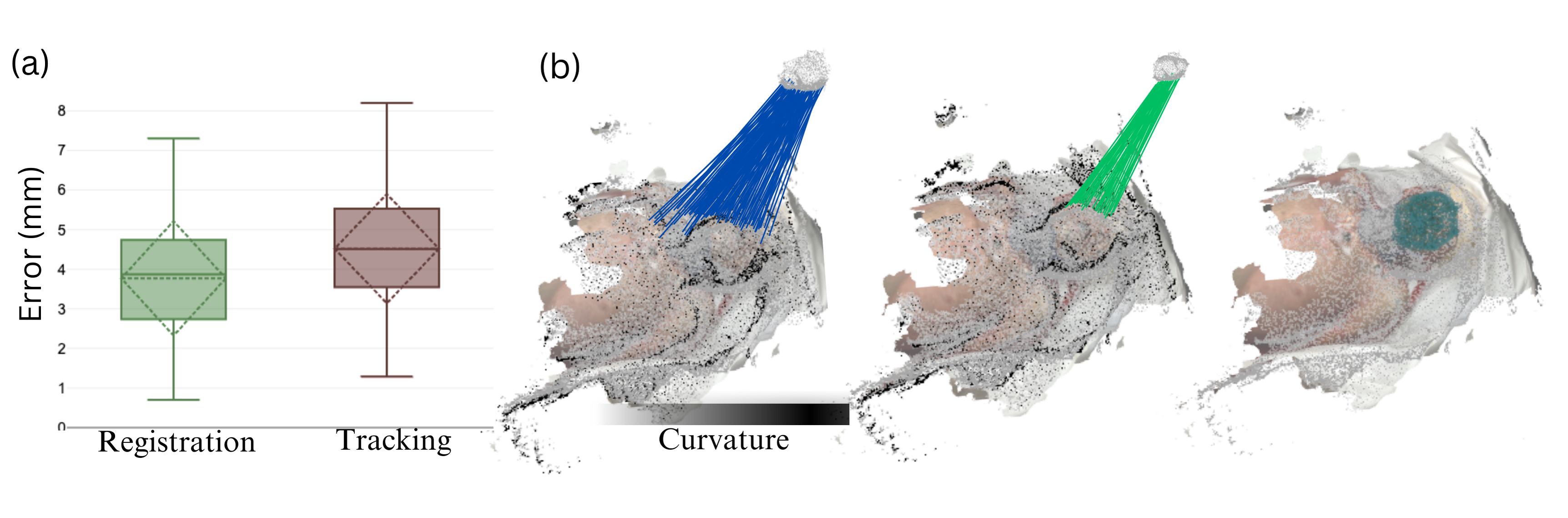}
    \caption{\textbf{(a)} Comparison of user study results. \textbf{(b)} Successful localization of the target anatomy (head) on AHAT sensor data. The head point cloud was generated from a segmentation of a full-body 3D scan and reconstruction, shown using a color-coded underlying representation. Initial FPFH-based correspondences are visualized in blue, inlier correspondences identified by the registration algorithm are shown in green, and the final localized target region is highlighted in green.}
    \label{fig:userstudy}
    \end{centering}
\end{figure}

\subsection{User Study and Cadaveric Use Case} 
To evaluate the real-time performance of our method, we conducted a user study with 12 participants. As illustrated in \cref{fig:teaser} (a), a virtual head model was registered to a 3D-printed head phantom (fabricated using the Bambu Lab X1C printer), with the peripheral features of the head deliberately occluded. Virtual targets were rendered as holographic dots over the registered model, each corresponding to a physical dot printed on the phantom (flat and colored). Since the holographic overlay occluded the participants’ view of the underlying physical model, participants were effectively blinded to the real target positions. An OptiTrack V120:Trio motion capture system (sub-millimeter accuracy) was used to track a probe with a localization marker. Eye calibration was performed individually for each participant to ensure accurate alignment. Each participant used their dominant hand to maneuver the probe tip to each virtual target and used their other hand to mark the position with an ultra-fine marker. Participants were allowed to reattempt points if they were not satisfied with their initial marking. A digital caliper was used to measure the linear deviation between the marked position and the corresponding physical dot, using the centroids of each as the measurement reference. A total of 25 points were labeled per participant during the initial registration condition.

After the initial session, the head phantom was repositioned to a new location to simulate patient movement, and the tracking module was activated. Participants then repeated the marking task for an additional 20 points. A within-subject analysis of variance (ANOVA) was used to compare the registration and tracking error distributions across conditions. The results showed no statistically significant difference between the registration module (Mean = 3.77 mm, SD = 1.45 mm) and the tracking module (Mean = 4.82 mm, SD = 1.38 mm), though a mean difference greater than 1 mm was observed.

To further demonstrate real-world applicability, we evaluated the performance of our coarse registration pipeline in a cadaveric use case. This component forms the foundational basis of both the registration and tracking modules; failure at this stage prevents any downstream registration or tracking. We tested TEASER++ with our improved correspondence generation strategy on cadaver head data. Initial correspondences were generated using FPFH descriptors, followed by TEASER++ for outlier-robust global registration. The target head point cloud was reconstructed from a 3D scan of the cadaver using a commercial 3D scanning app on iOS. Both the object and scene point clouds were uniformly downsampled to a similar density (approximately 10\% of original size). The results of this cadaver-based validation are shown in \cref{fig:userstudy} (b). Out of 12 attempts, TEASER++ achieved a 100\% success rate in global registration, with each registration
completed in under one second.

\section{Discussion}
This work presents a fully markerless registration and tracking framework that utilizes the native ToF sensor of the HoloLens 2 for surgical guidance. Our approach introduces a modular pipeline that delivers high registration accuracy and robust real-time tracking without relying on external cameras or physical fiducials. The design and evaluation of this system contribute key insights into the practical application of AR in intraoperative environments.

Our comprehensive evaluation of the AHAT depth sensor accuracy, followed by the implementation of a region-specific error correction strategy, offers a practical yet effective solution for mitigating localized sensor inaccuracies inherent to the HoloLens 2's ToF system. As detailed in \cref{sec:errorcorrection}, this correction approach is grounded in region–specific calibration rather than global compensation within one camera pose, allowing for higher-fidelity surface data in critical areas. We validated the effectiveness of this method through rigorous statistical analyses, including non-parametric testing of cadaveric measurements across multiple anatomical regions. The results confirmed that depth sensor error varies significantly between regions and that applying our localized correction method substantially reduces registration error, thereby enhancing the reliability of subsequent registration and tracking procedures.

In the development of our registration and tracking pipeline, we conducted a quantitative evaluation of 12 local registration algorithms using a custom benchmark dataset derived from the NIH3D collection \cref{sec:customdata}. This dataset was designed to closely represent real-world anatomical reconstructions obtained from preoperative MRI and CT imaging, with simulated sensor conditions introduced through the addition of Gaussian noise and partial data reduction. Each method was tested under varying conditions of point cloud overlap and relative rotation angles, and the RMSE was computed for each setting to assess performance. To systematically compare the suitability of these algorithms for different clinical tasks, we further introduced a custom scoring framework that modeled both registration and tracking performance. This comprehensive evaluation not only yielded valuable insights into the robustness and limitations of existing local registration methods in the context of medical point cloud registration, but also directly informed our algorithm selection. Specifically, the results guided the integration of a highly accurate and robust method into our registration module, and a computationally efficient variant into our tracking module. The strong performance of both modules, demonstrated in subsequent experiments, can be attributed in part to this rigorous algorithm selection process.

The evaluation of our registration module demonstrated a clear advantage over existing commercial SDKs in terms of accuracy across multiple translational and rotational motion scenarios. In particular, the use of curvature-aware feature sampling proved critical in achieving high precision. Compared to RGB-based methods, our ToF-based approach performed better registration, especially in tasks requiring larger translational alignment. Results show the value of depth sensing for medical registration tasks, especially when patient movement is minimal or highly localized.

Interestingly, although our registration module outperformed others in translational conditions, it exhibited slightly reduced performance under large vertical-axis rotations. We attribute this to limitations in the rotational invariance of FPFH descriptors and the degradation of the TEASER++ solver under significant angular displacement. These findings suggest a potential future direction: the integration of rotation-invariant descriptors or learned features that can more robustly handle viewpoint changes.

In terms of tracking, our system demonstrated stable and accurate pose estimation under real-time conditions, mostly maintaining sub-5~mm error across diverse movements. The tracking module, initialized by a precise registration result, showed runtime feasibility for surgical use. While the native runtime of 210~ms per frame yields a relatively low FPS rate, our use of visual interpolation to simulate 30~FPS addresses the smoothness requirement for user experience. Future work could explore further algorithmic optimization to improve frame rate without compromising accuracy.

The user study further validated the system’s ability to provide consistent guidance across participants. Importantly, although the registration module presents higher accuracy, the lack of statistically significant difference between registration and tracking phases suggests that \textit{EasyREG} can maintain target alignment even under motion, supporting dynamic surgical workflows.

Finally, the cadaver-based experiments highlight the robustness of the registration module in real-world anatomical contexts. In the presence of textureless surfaces and significant outliers, the system achieves successful global alignment. However, it is worth noting that the results presented here were obtained on a single HoloLens 2 device with one cadaver experiment. To generalize these findings, further studies are necessary. Although \textit{EasyREG} has demonstrated robustness under peripheral occlusion and real-world scenarios, further improvements tailored to handling non-rigid deformation and high-occlusion scenarios are promising directions for future research.

\section{Conclusion}
We presented \textit{EasyREG}, a fully markerless registration and tracking framework designed for surgical guidance using only the built-in sensors of the HoloLens 2. Our system integrates region-specific depth error correction, robust outlier-resistant registration, and real-time 6DoF tracking into a single, modular pipeline. Across controlled experiments, the registration module achieved a relative error of approximately 1.2 mm under lateral translation and 1.7 mm under depth translation. An approximately 1.8$^{\circ}$ error was achieved under vertical rotation. In a user study involving 12 participants, the system achieved a mean localization error of 3.77~mm during registration and 4.82~mm under dynamic tracking. A cadaveric experiment further validated system performance under real-life use cases. These results demonstrate that \textit{EasyREG} provides surgical level accuracy without requiring external hardware, and adaptability to various surgeries. The proposed framework enhances the practicality of AR navigation in surgery and offers insights for future work in non-rigid and occlusion-robust registration.

\acknowledgments{
The authors wish to thank A, B, C. 
}

\bibliographystyle{abbrv-doi-hyperref-narrow}

\bibliography{template}

\begin{thebibliography}{10}
\renewcommand*{\sfdefault}{PTSansNarrow-TLF}

\bibitem{ahmad2024automatic}
F.~Ahmad, J.~Xiong, and Z.~Xia.
\newblock Automatic feature-based markerless calibration and navigation method for augmented reality assisted dental treatment.
\newblock {\em IET Cyber-Systems and Robotics}, 6(4):e70003, 2024.

\bibitem{anandan2024surgical}
K.~Anandan, S.~P. Kumar, J.~J. Elona, D.~Balathay, T.~R. Ragav, and S.~Ganesan.
\newblock Surgical tool tracking: Comparative analysis of ar camera, optitrack ir, and realsense depth camera systems.
\newblock In {\em International Conference on Extended Reality}, pp. 163--177. Springer, 2024.

\bibitem{andress2018fly}
S.~Andress, A.~Johnson, M.~Unberath, A.~F. Winkler, K.~Yu, J.~Fotouhi, S.~Weidert, G.~Osgood, and N.~Navab.
\newblock On-the-fly augmented reality for orthopedic surgery using a multimodal fiducial.
\newblock {\em Journal of Medical Imaging}, 5(2):021209--021209, 2018.

\bibitem{aoki2019pointnetlk}
Y.~Aoki, H.~Goforth, R.~A. Srivatsan, and S.~Lucey.
\newblock Pointnetlk: Robust \& efficient point cloud registration using pointnet.
\newblock In {\em Proceedings of the IEEE/CVF conference on computer vision and pattern recognition}, pp. 7163--7172, 2019.

\bibitem{barcali2022augmented}
E.~Barcali, E.~Iadanza, L.~Manetti, P.~Francia, C.~Nardi, and L.~Bocchi.
\newblock Augmented reality in surgery: a scoping review.
\newblock {\em Applied Sciences}, 12(14):6890, 2022.

\bibitem{birlo2022utility}
M.~Birlo, P.~E. Edwards, M.~Clarkson, and D.~Stoyanov.
\newblock Utility of optical see-through head mounted displays in augmented reality-assisted surgery: A systematic review.
\newblock {\em Medical Image Analysis}, 77:102361, 2022.

\bibitem{biswas2023recent}
P.~Biswas, S.~Sikander, and P.~Kulkarni.
\newblock Recent advances in robot-assisted surgical systems.
\newblock {\em Biomedical Engineering Advances}, 6:100109, 2023.

\bibitem{bouaziz2013sparse}
S.~Bouaziz, A.~Tagliasacchi, and M.~Pauly.
\newblock Sparse iterative closest point.
\newblock In {\em Computer graphics forum}, vol.~32, pp. 113--123. Wiley Online Library, 2013.

\bibitem{canton2024feasibility}
S.~P. Canton, C.~N. Austin, F.~Steuer, S.~Dadi, N.~Sharma, N.~M. Kass, D.~Fogg, E.~Clayton, O.~Cunningham, D.~Scott, et~al.
\newblock Feasibility and usability of augmented reality technology in the orthopaedic operating room.
\newblock {\em Current Reviews in Musculoskeletal Medicine}, 17(5):117--128, 2024.

\bibitem{choy2020deep}
C.~Choy, W.~Dong, and V.~Koltun.
\newblock Deep global registration.
\newblock In {\em Proceedings of the IEEE/CVF conference on computer vision and pattern recognition}, pp. 2514--2523, 2020.

\bibitem{cutolo2021device}
F.~Cutolo, N.~Cattari, M.~Carbone, R.~D’Amato, and V.~Ferrari.
\newblock Device-agnostic augmented reality rendering pipeline for ar in medicine.
\newblock In {\em 2021 IEEE International Symposium on Mixed and Augmented Reality Adjunct (ISMAR-Adjunct)}, pp. 340--345. IEEE, 2021.

\bibitem{de2024integrating}
M.~De~Jesus Encarnacion~Ramirez, G.~Chmutin, R.~Nurmukhametov, G.~R. Soto, S.~Kannan, G.~Piavchenko, V.~Nikolenko, I.~E. Efe, A.~R. Romero, J.~N. Mukengeshay, et~al.
\newblock Integrating augmented reality in spine surgery: redefining precision with new technologies.
\newblock {\em Brain Sciences}, 14(7):645, 2024.

\bibitem{duan2025localization}
H.~Duan, Y.~Yang, W.~L. Niu, D.~Anders, A.~M. Dreisbach, D.~Holley, B.~L. Franc, S.~L. Perkins, C.~Leuze, B.~L. Daniel, et~al.
\newblock Localization of sentinel lymph nodes using augmented-reality system: a cadaveric feasibility study.
\newblock {\em European Journal of Nuclear Medicine and Molecular Imaging}, pp. 1--10, 2025.

\bibitem{elfring2010assessment}
R.~Elfring, M.~de~la Fuente, and K.~Radermacher.
\newblock Assessment of optical localizer accuracy for computer aided surgery systems.
\newblock {\em Computer Aided Surgery}, 15(1-3):1--12, 2010.

\bibitem{evans2025augmented}
M.~Evans, S.~Kang, A.~Bajaber, K.~Gordon, and C.~Martin~III.
\newblock Augmented reality for surgical navigation: A review of advanced needle guidance systems for percutaneous tumor ablation.
\newblock {\em Radiology: Imaging Cancer}, 7(1):e230154, 2025.

\bibitem{furtado2019comparative}
J.~S. Furtado, H.~H. Liu, G.~Lai, H.~Lacheray, and J.~Desouza-Coelho.
\newblock Comparative analysis of optitrack motion capture systems.
\newblock In {\em Advances in Motion Sensing and Control for Robotic Applications: Selected Papers from the Symposium on Mechatronics, Robotics, and Control (SMRC’18)-CSME International Congress 2018, May 27-30, 2018 Toronto, Canada}, pp. 15--31. Springer, 2019.

\bibitem{gao2019filterreg}
W.~Gao and R.~Tedrake.
\newblock Filterreg: Robust and efficient probabilistic point-set registration using gaussian filter and twist parameterization.
\newblock In {\em Proceedings of the IEEE/CVF conference on computer vision and pattern recognition}, pp. 11095--11104, 2019.

\bibitem{groenenberg2024feasibility}
A.~Groenenberg, L.~Brouwers, M.~Bemelman, T.~J. Maal, J.~M. Heyligers, and M.~M. Louwerse.
\newblock Feasibility and accuracy of a real-time depth-based markerless navigation method for hologram-guided surgery.
\newblock {\em BMC Digital Health}, 2(1):11, 2024.

\bibitem{hersh2021augmented}
A.~Hersh, S.~Mahapatra, C.~Weber-Levine, T.~Awosika, J.~N. Theodore, H.~M. Zakaria, A.~Liu, T.~F. Witham, and N.~Theodore.
\newblock Augmented reality in spine surgery: a narrative review.
\newblock {\em HSS Journal{\textregistered}}, 17(3):351--358, 2021.

\bibitem{hu2024artificial}
X.~Hu, F.~Cutolo, H.~Iqbal, J.~Henckel, and F.~R. y~Baena.
\newblock Artificial intelligence-driven framework for augmented reality markerless navigation in knee surgery.
\newblock {\em IEEE Transactions on Artificial Intelligence}, 2024.

\bibitem{jian2010robust}
B.~Jian and B.~C. Vemuri.
\newblock Robust point set registration using gaussian mixture models.
\newblock {\em IEEE transactions on pattern analysis and machine intelligence}, 33(8):1633--1645, 2010.

\bibitem{khan2015factors}
D.~Khan, S.~Ullah, and I.~Rabbi.
\newblock Factors affecting the design and tracking of artoolkit markers.
\newblock {\em Computer Standards \& Interfaces}, 41:56--66, 2015.

\bibitem{kihara2024evaluating}
T.~Kihara, A.~Keller, T.~Ogawa, M.~Armand, and A.~Martin-Gomez.
\newblock Evaluating the feasibility of using augmented reality for tooth preparation.
\newblock {\em Journal of Dentistry}, 148:105217, 2024.

\bibitem{kim2024application}
Y.~C. Kim, C.-U. Park, S.~J. Lee, W.~S. Jeong, S.~W. Na, and J.~W. Choi.
\newblock Application of augmented reality using automatic markerless registration for facial plastic and reconstructive surgery.
\newblock {\em Journal of Cranio-Maxillofacial Surgery}, 52(2):246--251, 2024.

\bibitem{li2024evd}
H.~Li, W.~Yan, D.~Liu, L.~Qian, Y.~Yang, Y.~Liu, Z.~Zhao, H.~Ding, and G.~Wang.
\newblock Evd surgical guidance with retro-reflective tool tracking and spatial reconstruction using head-mounted augmented reality device.
\newblock {\em IEEE Transactions on Visualization and Computer Graphics}, 2024.

\bibitem{li2024navigate}
H.~Li, W.~Yan, J.~Zhao, Y.~Ji, L.~Qian, H.~Ding, Z.~Zhao, and G.~Wang.
\newblock Navigate biopsy with ultrasound under augmented reality device: Towards higher system performance.
\newblock {\em Computers in Biology and Medicine}, 174:108453, 2024.

\bibitem{lin2021preliminary}
L.~Lin, C.~Xu, Y.~Shi, C.~Zhou, M.~Zhu, G.~Chai, and L.~Xie.
\newblock Preliminary clinical experience of robot-assisted surgery in treatment with genioplasty.
\newblock {\em Scientific Reports}, 11(1):6365, 2021.

\bibitem{low2004linear}
K.-L. Low.
\newblock Linear least-squares optimization for point-to-plane icp surface registration.
\newblock {\em Chapel Hill, University of North Carolina}, 4(10):1--3, 2004.

\bibitem{lv2023kss}
C.~Lv, W.~Lin, and B.~Zhao.
\newblock Kss-icp: point cloud registration based on kendall shape space.
\newblock {\em IEEE Transactions on image processing}, 32:1681--1693, 2023.

\bibitem{ma2019augmented}
L.~Ma, W.~Jiang, B.~Zhang, X.~Qu, G.~Ning, X.~Zhang, and H.~Liao.
\newblock Augmented reality surgical navigation with accurate cbct-patient registration for dental implant placement.
\newblock {\em Medical \& biological engineering \& computing}, 57:47--57, 2019.

\bibitem{malhotra2023augmented}
S.~Malhotra, O.~Halabi, S.~P. Dakua, J.~Padhan, S.~Paul, and W.~Palliyali.
\newblock Augmented reality in surgical navigation: a review of evaluation and validation metrics.
\newblock {\em Applied Sciences}, 13(3):1629, 2023.

\bibitem{martin2023sttar}
A.~Martin-Gomez, H.~Li, T.~Song, S.~Yang, G.~Wang, H.~Ding, N.~Navab, Z.~Zhao, and M.~Armand.
\newblock Sttar: surgical tool tracking using off-the-shelf augmented reality head-mounted displays.
\newblock {\em IEEE Transactions on Visualization and Computer Graphics}, 2023.

\bibitem{moreta2022evaluation}
R.~Moreta-Mart{\'\i}nez, I.~Rubio-P{\'e}rez, M.~Garc{\'\i}a-Sevilla, L.~Garc{\'\i}a-Elcano, and J.~Pascau.
\newblock Evaluation of optical tracking and augmented reality for needle navigation in sacral nerve stimulation.
\newblock {\em Computer methods and programs in biomedicine}, 224:106991, 2022.

\bibitem{NIH3D}
NIH.
\newblock Nih 3d model database.
\newblock \url{https://3d.nih.gov/}.
\newblock Accessed: 2024-09-30.

\bibitem{PCL}
PointCloudLibrary.
\newblock Point cloud normal generation.
\newblock \url{https://pointclouds.org/}, 2011.
\newblock Accessed: 2024-10-30.

\bibitem{sarode2019pcrnet}
V.~Sarode, X.~Li, H.~Goforth, Y.~Aoki, R.~A. Srivatsan, S.~Lucey, and H.~Choset.
\newblock Pcrnet: Point cloud registration network using pointnet encoding.
\newblock {\em arXiv preprint arXiv:1908.07906}, 2019.

\bibitem{suenaga2015vision}
H.~Suenaga, H.~H. Tran, H.~Liao, K.~Masamune, T.~Dohi, K.~Hoshi, and T.~Takato.
\newblock Vision-based markerless registration using stereo vision and an augmented reality surgical navigation system: a pilot study.
\newblock {\em BMC medical imaging}, 15:1--11, 2015.

\bibitem{tadros2019ductal}
A.~B. Tadros, B.~D. Smith, Y.~Shen, H.~Lin, S.~Krishnamurthy, A.~Lucci, C.~H. Barcenas, R.~F. Hwang, G.~Rauch, L.~Santiago, et~al.
\newblock Ductal carcinoma in situ and margins< 2 mm: contemporary outcomes with breast conservation.
\newblock {\em Annals of surgery}, 269(1):150--157, 2019.

\bibitem{tondin2022evaluation}
G.~M. Tondin, M.~d. O. C.~D. Leal, S.~T. Costa, R.~Grillo, C.~R.~P. Jodas, and R.~G. Teixeira.
\newblock Evaluation of the accuracy of virtual planning in bimaxillary orthognathic surgery: a systematic review.
\newblock {\em British Journal of Oral and Maxillofacial Surgery}, 60(4):412--421, 2022.

\bibitem{wang2019practical}
J.~Wang, Y.~Shen, and S.~Yang.
\newblock A practical marker-less image registration method for augmented reality oral and maxillofacial surgery.
\newblock {\em International journal of computer assisted radiology and surgery}, 14:763--773, 2019.

\bibitem{wang2020unsupervised}
L.~Wang, X.~Li, and Y.~Fang.
\newblock Unsupervised learning of 3d point set registration.
\newblock {\em arXiv preprint arXiv:2006.06200}, 2020.

\bibitem{wang2019deep}
Y.~Wang and J.~M. Solomon.
\newblock Deep closest point: Learning representations for point cloud registration.
\newblock In {\em Proceedings of the IEEE/CVF international conference on computer vision}, pp. 3523--3532, 2019.

\bibitem{wang2019prnet}
Y.~Wang and J.~M. Solomon.
\newblock Prnet: Self-supervised learning for partial-to-partial registration.
\newblock {\em Advances in neural information processing systems}, 32, 2019.

\bibitem{yang2015globally}
J.~Yang, H.~Li, D.~Campbell, and Y.~J. Go-ICP.
\newblock A globally optimal solution to 3d icp point-set registration., 2015, 38.
\newblock {\em DOI: https://doi. org/10.1109/TPAMI}, pp. 2241--2254, 2015.

\bibitem{zhang2021fast}
J.~Zhang, Y.~Yao, and B.~Deng.
\newblock Fast and robust iterative closest point.
\newblock {\em IEEE Transactions on Pattern Analysis and Machine Intelligence}, 44(7):3450--3466, 2021.

\end{thebibliography}
\clearpage
\listofchanges
\end{document}